**How does spatial structure affect psychological restoration? A method based on Graph Neural Networks and Street View Imagery**

Haoran Ma[a,b], Yan Zhang[c], Pengyuan Liu[d], Fan Zhang[e], Pengyu Zhu[a*]

[a] *Division of Public Policy, Hong Kong University of Science and Technology, Hong Kong, China;*

[b] *School of Design, Jiangnan University, Wuxi, China;*

[c] *Institute of Space and Earth Information Science, The Chinese University of Hong Kong, Hong Kong, China;*

[d] *School of Geographical Sciences, Nanjing University of Information Science and Technology, China;*

[e] *Institute of Remote Sensing and Geographical Information System, School of Earth and Space Sciences, Peking University, China;*

**ABSTRACT:** The Attention Restoration Theory (ART) presents a theoretical framework with four essential indicators (being away, extent, fascinating, and compatibility) for comprehending urban and natural restoration quality. However, previous studies relied on non-sequential data and non-spatial dependent methods, which overlooks the impact of spatial structure—defined here as the positional relationships between scene entities—on restoration quality. The past methods also make it challenging to measure restoration quality on an urban scale. In this work, a spatial-dependent graph neural networks (GNNs) approach is proposed to reveal the relation between spatial structure and restoration quality on an urban scale. Specifically, we constructed two different types of graphs at the street and city levels. The street-level graphs, using sequential street view images (SVIs) of road segments to capture position relationships between entities, were used to represent spatial structure. The city-level graph, modeling the topological relationships of roads as non-Euclidean data structures and embedding urban features (including Perception-features, Spatial-features, and Socioeconomic-features), was used to measure restoration quality. The results demonstrate that: 1) spatial-dependent GNNs model outperforms traditional methods (Acc = 0.735, F1 = 0.732); 2) spatial structure portrayed through sequential SVIs data significantly influences restoration quality; 3) spaces with the same restoration quality exhibited distinct spatial structures patterns. This study clarifies the association between spatial structure and restoration quality, providing a new perspective to improve urban well-being in the future.

restoration theory

1. Introduction

The urban landscape, home to over half the world's population, is in a state of flux and is predicted to accommodate 75% of the global populace by 2050 (Ritchie and Roser,2018). This rapid urbanization necessitates a focus on mitigating the adverse impacts of the urban environment on human health. It is particularly crucial to promote physical health, manage stress, and prevent stress-related diseases (Akpınar,2021; Cetin et al.,2021; Liu et al.,2020). Urban design emerges as a broad-based solution, recognized for its potential to address these mental health issues (UNFPA, 2007; Keniger et al., 2013; Hough,2014).

In contrast to rural life, urban life has been associated with poorer mental health, as evidenced in numerous studies (Peen et al.,2010). It has been confirmed that exposure to forests and green spaces is beneficial for human mental health, which effectively reduces stress (Capaldi et al.,2014), improves mood (Berman et al.,2008), and restores depleted cognitive resources (Akpınar,2021; Liu et al.,2020), as supported by the Attention Restoration Theory (ART) (Kaplan,1995; Hartig et al.,1991). Nevertheless, when compared to nature, these studies frequently drive urban areas into unequal places and show that they have few restoration qualities, undermining the psychological restorative potential of urban spaces (Kaplan and Berman,2010; Schertz and Berman,2019). Although some researchers have opposite perspectives and confirmed that restoration exists in urban spaces, they are usually limited to specific locations (such as waterfront) and have not fully considered the spatial structure and geographical relevance of urban spaces influence restorative quality (Burmil et al.,1999; Lindal and Hartig,2013).

Spatial structure can be understood as the positional relationship between the physical entities. It has been confirmed that they are associated with the layout of streets (Ashihara, 1986), landscape design (Cullen, 2012), and land use function (Zube,1987) in past studies, which in turn impact human perception (Lynch,1984). However, it is unclear how spatial structures impact psychological restoration. Celikors and Wells (2022) showed that two images with similar visual elements could elicit different degrees of psychological restoration due to the change in their spatial structure. Revealing the effect of spatial structure on psychological restoration can potentially enrich space interpretability. Besides this, the intrinsic association between spatial structure is apparent, as the city is a spatial continuum (Horton and Reynolds, 1971; Carmona,2014,2021). According to the first law of geography, similar things are more relevant (Tobler,1970). For example, waterfront spaces will affect the surrounding landscape design and spatial structure, which will further affect their restorative quality (Burmil et al.,1999). However, due to the limitations of non-sequential data sources and non-spatial dependent methods, which struggle to represent spatial structure and measure restoration quality on an urban scale (Nordh et al., 2009; Lindal and Hartig, 2013), it becomes imperative to re-examine the relationship between the spatial structure of urban spaces and their restorative qualities. This approach will further assist researchers in uncovering distribution patterns and spatial relevance of restoration.

Street View images (SVIs) have high spatial resolution and provide sequential data (interval of 50 meters) within a wealth of urban information, which has been widely used in the study of the urban form (Gong et al., 2019; Ito and Biljecki, 2021), visual perception (Biljecki and Ito, 2021), and health behavior (Fan et al., 2023; Rzotkiewicz et al., 2018). In recent years, the use of such data to assess urban restoration quality has garnered increasing attention. Some studies have explored the potential of SVIs to predict urban restoration quality at the city level and demonstrated efficiency and accuracy (Han et al.,2023; Ma et al.,2023). However, only non-sequential images (single images) and their visual information were taken into account, with the influence of spatial structure and geographic relevance being disregarded. Specifically, they overlooked the intrinsic and extrinsic connections of spatial entities within the SVIs (Zhang et al.,2023; Liang et al.,2023).

Graph neural networks (GNNs) have shown serious advantages in capturing the relation of everything and predicting its attributes, which are frequently used in traffic flow, urban population movement, and social perception (Liu and Biljecki, 2022; Zhang et al., 2023; Kipf and Welling, 2016). Inspired by these studies, we seek to use graphs to represent urban spatial structures and measure restoration quality. Specifically, in this study, we proposed a spatial-dependent GNNs approach to reveal the relation between spatial structure and restoration quality on an urban scale, which involved two different types of graphs - street-level graph and city-level graph, respectively. The spatial structure was represented by street-level graphs, which used sequential street view images (SVIs) of road segments to capture position relationships between entities. To measure restoration quality, the city-level graph, which modeled the topological relationships of roads as non-Euclidean data structures and integrated urban features (including Socioeconomic-features, Perception-features, and Spatial-features), was utilized. To the best of our knowledge, using this method that we proposed to assess the urban restoration quality and reveal the spatial structure effect is groundbreaking.

The present study contributes significantly to the research landscape in three key ways:

• We propose a spatial-dependent GNN method for measuring urban restoration quality, which can effectively predict the restoration quality on a city-level graph by capturing spatial dependencies based on road topology.

• We explore the influence of spatial structure on the quality of urban restoration, verifying that street-level graphs accurately capture the intrinsic and extrinsic connections between entities through sequential SVIs.

• We uncover the spatial structure distinct among high-quality restoration spaces, and identified four different categories of spatial structures based on cluster analysis enhancing their interpretability.

**2. Related works**

**2.1 Restorative quality in urban environments**

Attention Restoration Theory (ART) provided a framework for understanding the mental health benefits of environmental interactions. The psychological recovery and improvement of cognitive functioning after mental fatigue were referred to as *Restoration* in ART (Kaplan,1995). According to Kaplan and Kaplan (1989), to restore cognitive resources, the environment should possess the qualities of Being away, Extent, Fascination, and Compatibility. Being away allowed us to distance ourselves from the routine fatigue of daily life. Extent refers to the expansiveness of the environment and the degree to which it invites exploration. Fascination is characterized by an environment's ability to attract our interest without consuming our attentional resources. Compatibility is the degree of alignment between an individual's needs or preferences and the environment's characteristics.

The advantages of interacting with natural environments were evident, such as reducing anxiety (Felsten, 2009), alleviating stress (Capaldi et al., 2014), improving mood (Berman et al., 2008), and performance in tasks requiring attention and working memory (Bratman et al.,2015; Stenfors et al.,2019). However, previous studies have focused primarily on the restorative benefits of natural environments and often compare them with urban environments (Hartig et al., 1991; Ulrich et al., 1991). Urban environments, on the other hand, were generally believed to deplete mental and attentional resources (Kaplan and Berman,2010; Schertz and Berman,2019). Such categorization can exacerbate the perceived contrast between "natural" and "urban" environments. Yet, urban elements such as green-blue spaces (Li et al., 2023) and walkable spaces (Han et al., 2023) have been associated with mental restoration. Some urban locations also possessed restoration potentials, such as art galleries (Clow and Fredhoi,2006), shopping centers and cafes (Staats et al.,2016). However, these studies often treat urban environments as a homogeneous category, overlooking the variations in spatial structure and geographical relevance (Velarde et al.,2007).

Spatial elements can have different meanings in different contexts, and changes in spatial patterns result in variations in spatial structures. Historical studies and theories have suggested an association between human psychological perceptions and urban spatial structures (Lynch,1984; Ashihara,1986; Zube,1987). However, there is no documented evidence of a direct correlation between psychological restoration and spatial structures. Celikors and Wells (2022) proposed that similar visual properties could elicit different restoration judgments due to inherent spatial representation, highlighting the need for further investigation. Moreover, the lived experiences of city dwellers, which are shaped by the physical environment and are continuous, should be taken into account as the city is a spatial continuum and a system (Nordh et al.,2009; Lindal and Hartig,2013). Previous research often relied on non-sequential, non-geotagged data (Han et al.,2023; Ma et al., 2023), but similar and close things are more relevant in urban space, according to the first law of geography (Tobler, 1970). Thus, it's essential to consider both spatial structures and geographical relevance on an urban scale. However, traditional methods are often constrained by time and costs, focusing on specific locations and utilizing small sample sizes (Nordh et

al.,2009; Lindal and Hartig,2013). This necessitates proposing a new research framework and data source.

**2.2 Street view imagery in urban studies**

In recent years, crowd-sourced data have been widely applied in urban studies, such as Street View Images (SVIs) (Biljecki and Ito, 2021). It contains rich urban information and is extensively used to extract environment features (Tang and Long,2019; Zhou et al.,2019), build urban knowledge graphs (Zhang et al.,2023), analyze environmental health (Rzotkiewicz et al., 2018), and predict humans' perception (Zhang et al., 2018; Zhao et al., 2023). Zhang et al. (2018) pioneered the concept of establishing connections between SVIs and six human perceptions (namely, beautiful, boring, lively, depressing, wealthy, and safety) by calculating the visual perception of urban features through semantic segmentation. Subsequent research has confirmed the accuracy of this data in predicting perceptions of safety (Kang et al., 2023). Additionally, with the development of computer vision, such as image classification (Hu et al., 2020), semantic segmentation (Lauko et al.,2020), and object detection (Zhao et al.,2023), it has become efficient to analyze large-scale SVIs to investigate urban issues.

Moreover, SVIs have a sequential attribute and high spatial resolution, allowing researchers to study perceptual variations in cities on a large scale, in detail, and over time. For instance, Zhang et al. (2023) employed language information extracted from SVIs to capture the intrinsic and extrinsic relationship between scene entities to predict urban land functions. However, there have not been many cases in which data are used to evaluate the quality of urban restoration. In a recent study, Han et al. (2023) pioneered the exploration of a large-scale urban restoration quality assessment framework via 1,250 SVIs, achieving accurate results. Ma et al. (2023) also employed SVIs exploring the relationship between visual features and restoration quality on a campus scale. However, a significant limitation is that they only deal with non-sequential images (single images), which means that the relationship between scene entities on a road segment cannot be captured. Therefore, it is important to uncover the value of sequential attributes of SVIs, which entity relationships between streetscapes, represent spatial structures, and can further explore their relationship to restorative perception.

**2.3 Spatially dependent graph neural networks**

Graph Neural Networks (GNNs) have been shown to outperform traditional models in several areas, such as traffic flow prediction, urban population mobility, and social sensing (Liu and Biljecki, 2022). They are capable of handling structured non-Euclidean data, extracting spatial features from graphs for efficient learning (Zhang et al.,2021; Yao et al.,2021), and capturing the spatial dependency and heterogeneity of urban features (Liu et al., 2023; Liang et al., 2023). According to the first law of geography, city regions within a specific range may become increasingly similar due to the strong correlation between urban scenes and their neighboring areas (Tobler,1970). Previous studies have found that the spatial relationship between neighbors, represented by graph theory, can identify high-level features. Zhang et al. (2023) constructed a city knowledge graph containing urban geographic information using SVIs and verified the feasibility and accuracy of GNNs in representing urban spatial structures.

GNNs can learn the deep representation of spatial relationships between adjacent scenes through aggregation algorithms, where each node can aggregate features from its neighbors (Defferrard et al., 2016). Based on the foundational principles of GNNs, several derived models have been developed, such as Graph Convolutional Networks (GCN) (Kipf and Welling, 2016), Graph Attention Networks (GAT) (Veličković et al., 2017), and GraphSAGE (Hamilton et al., 2017). Thanks to the powerful data organization capability and the ability to handle non-Euclidean data structures, GNNs can integrate various modalities of urban data into graph neural networks for downstream tasks. Examples of such data include SVIs (Liu et al., 2023), Points of Interest (POI) (Xu et al., 2022a), land use (Liang et al., 2023), and social media data (Liu and De Sabbata, 2021), achieving state-of-the-art performance. For instance, Xu et al. (2022a) combined visual features of cities with POI data for urban scene classification, improving the precision by 13% compared to traditional methods. Although GNNs have the powerful ability to handle various urban tasks, there has been no research applying them to the study of the quality of urban recovery. Thus, our study is pioneering.

## 3. Methods and data

The research framework is established with five parts (Figure 1): 1) Extracting features and embedding spatial structure (street-level graph). 2) Feature aggregation and graph construction (city-level graph). 3) Data labeling and enhancement. 4) Model training and evaluation. 5) Overall analysis. This section will provide a detailed introduction to the 1-4 parts, and part 5 will be analyzed in detail in the results section. Our research area is within the third ring road of Wuhan (Figure A8). For this study, we collected four types of data: the OpenStreetMap (OSM) road network consisting of 5,075 roads, 64,750 panoramic SVIs from Baidu Maps, as well as Points of Interest (POI), and housing price data. All data were collected in June 2023.

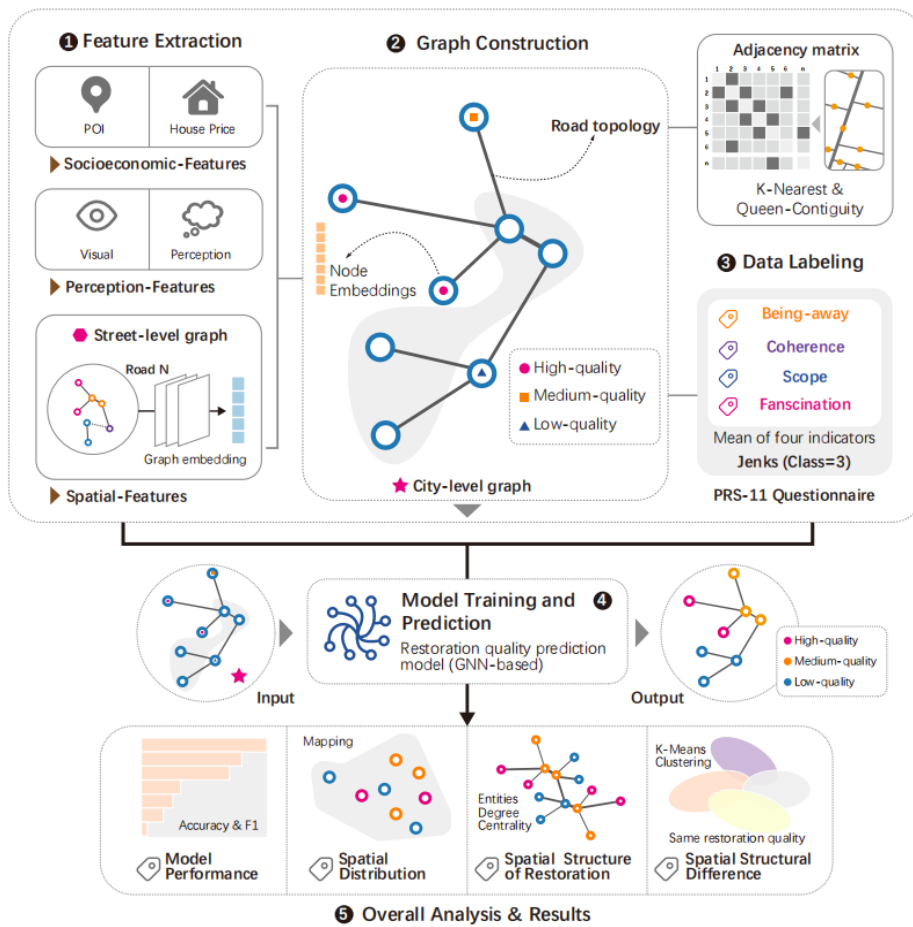

**Figure 1:** Research framework. Our research is divided into 5 parts: 1). Feature extraction, this part extracts three levels of urban features: socioeconomic, perceptional, and spatial; 2). Graph construction, this part focuses on aggregating urban features and constructing a city-level graph through road topology; 3). Data labeling, this part involves using the PRS-11 questionnaire to score and label data; 4). Model training and prediction, this part treats the city-level graph as input to predict urban restoration quality through GNN-based models; 5). Overall analysis and results, there are four results represented.

### 3.1 Extracting features and embedding spatial structure

Urban features were categorized into three classes, which were derived from multiple sources of urban: 1) Perception-features (SVIs were utilized to deduce perception features, which included object quantity, proportion, and perceptual scores), 2) Spatial-features (SVIs were utilized to construct street-level graphs and obtain these embedded features), and 3) Socioeconomic-features (originating from POI and housing prices data). Table S1 shows the results of data normalization for all variables.

**Perception-features.** Urban space quality evaluations often extract physical components from SVIs, such as enclosure, greenery, openness, and safety (Tang and Long, 2019; Zhou et al., 2019). Previous studies have indicated that shallow visual features (pixel level) and deep visual features (semantic and object level) of images can affect perceptions of attention restoration (Ibarra et al.,2017; Celikors and Wells,2022; Valtchanov and Ellard,2015). We used OpenCV to calculate the pixel-level information of street view images (Zhao et al.,2023). For the extraction of deep visual features, including semantic segmentation at the semantic level to compute the proportion of physical elements and object detection at the object level to calculate the number of entities, we employed MaskFomer (Cheng et al.,2022) and DETR (Carion et al.,2020). MaskFomer is capable of segmenting 150 object categories, and DETR can detect 90 object categories, which have state-of-the-art performance. We used these models to extract visual features from SVIs.

Perceptions of cities also have an impact on the restoration of attention. More aesthetically appealing places tend to be more attractive, aligning with Kaplan's concept of a fascinating space that can restore attention resources (Berman et al., 2008). For the evaluation of perceptual scores, we used the Place Pulse 2.0 dataset, which includes 110,988 images from 56 cities in 28 countries, with six labels: depressing, boring, beautiful, safe, lively, and wealthy (Dubey et al.,2016). Based on examining its initial version, no significant cultural or individual preference biases were found, indicating its feasibility in global research (Salesses et al.,2013). We used a pre-trained model provided by Yao et al. (2019) to predict the perceptual scores of 64,750 SVIs in Wuhan city. Table 1 summarizes the models and algorithms used for feature extraction.

**Table 1**
Summary of feature extraction models and algorithms.

| Features | Model | Dateset | Variables |
|---|---|---|---|
| Pixel-level | OpenCV | - | 5 categories |
| Object-level | DETR (Carion et al., 2020) | COCO 2017 | 90 categories |
| Semantic-level | MaskFomer (Cheng et al., 2022) | ADE20K | 150 categories |
| Perception | ResNet (Yao et al., 2019) | Place Pulse2.0 | 6 categories |

**Spatial-features (street-level graph).** As previously discussed, due to the intrinsic and extrinsic association in spatial relationships between entities, there is numerous geographic relevance between similar elements in different urban spaces (Kang et al., 2018; Zhang et al., 2023). To capture the relationships between urban scenes, we take multiple street views contained in each road segment as units (for instance, the same buildings can be identified in adjacent street views, establishing a relationship between the two images). As shown in Figure 2:

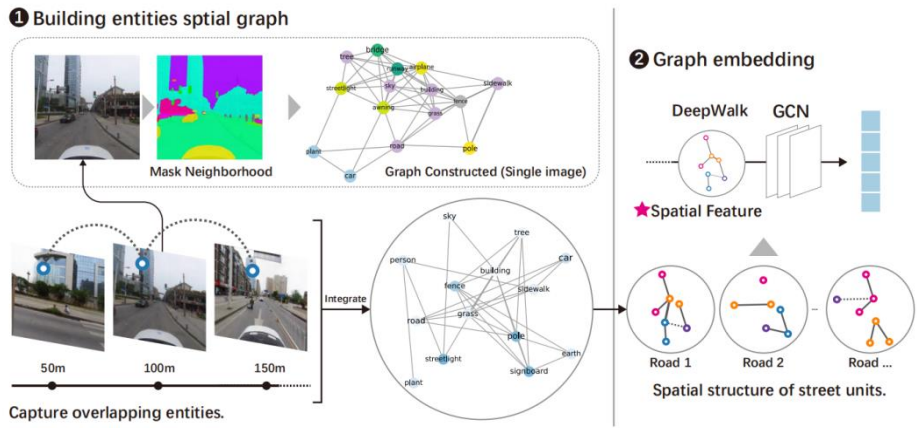

Figure 2: Embedding the spatial structure. First, constructing the spatial graph of entities by inferring the adjacency relationships between entities in multiple street views to represent the spatial structure of urban roads. Second, graph embedding. Treating road as the smallest units, using DeepWalk and GCN (two-layer convolution, the hidden layer computational unit is 32 and 16) to embed the street structure into feature vectors.

First, building entities spatial graph. We use MaskFomer to calculate entity classes $K = 150$ for each street view and calculate an adjacency threshold to determine the adjacency relationships between entities. Then, we construct an undirected graph $G = (V, E)$ with $K$ nodes, where $V$ represents the nodes set and $E$ represents the edges set. Specifically, for each classes $K$, we find all their pixel coordinates $(x_1, y_1), (x_2, y_2), \ldots, (x_n, y_n)$ with predicted values equal to $i$, calculate their centroid coordinates $c = [(x_1 + x_2 + \ldots + x_n), (y_1 + y_2 + \ldots + y_n)] \div n$, and add $(i, c_i)$ to the nodeset $V$. Then, for any two nodes, such as $(i, c_i)$ and $(j, c_j)$, if the Euclidean distance $dist(i, j)$ between them is less than a threshold $T$, that is, $dist(ci, cj) < T$, we add an $edge\ (i,j)$ to the edge set $E$. After repeated testing fifty times, the threshold $T$ in this study is set at 45 to accurately express the spatial adjacency relationship.

Second, graph embedding. Multiple street view images from the same street are integrated to represent the spatial structure of the smallest units. We use the DeepWalk algorithm to encode each node, where DeepWalk is used to generate sequence data related to the nodes in the graph (Perozzi et al.,2014). These sequence data are input to GCN (Kipf and Welling,2016) to generate a vector ($D = 5$) as an explicit representation of the spatial structure of the street. The code is shared on GitHub: https://github.com/MMHHRR/Restoration-Topology

**Socioeconomic-features.** In addition, we used POI data to examine service indicators and housing price data to examine the economic conditions of the regions, these features have been proven to be related to the quality of environmental restoration (Subiza-Pérez et al.,2021; Samus et al.,2022; Luttik,2000). We collected these data in June 2023 and used ArcGIS to calculate the average POI density and the average price of housing (yuan per square meter).

**3.2 Features aggregation and graph construction (city-level graph)**

Urban environments often display spatial variability, wherein neighboring spaces exhibit similarities (Tobler,1970). The study area consists of 5,075 OSM roads (road units), with the midpoint of each road serving as a node $N$ (5,075) on the undirected graph, because the road network shapes urban functions and traffic as the skeleton of the city (Hong and Yao, 2019). We aggregate the urban features mentioned in section 3.1into the roads. We set a 25m wide buffer based on OSM roads (50m can cover most urban road widths), which is spatially linked to the urban spatial feature data and mapped to the corresponding road segments. The dimension of the feature is $D = 279$.

Simultaneously, we employ the topology of the road network to create a spatial weight matrix, representing relationships between adjacent roads. Specifically, we use a matrix $n \times n$ ($n$ is the number of all road segments), and $A$ to express the adjacency relationships between the roads. If there is an adjacency relationship between streets $i$ and $j$, $A_{i,j}$ is assigned a value of one; otherwise, $A_{i,j}$ is assigned a value of zero. To depict the adjacency relationships of city streets, we adopt two approaches. First, we determine proximate roads using the weighting of the K-Nearest (K = 5), which signifies their adjacency relationships (25,375 neighboring relationships). This method prevents dangling roads but may also categorize certain non-intersecting roads as nearby (Zhu et al.,2020). The second approach is the Queen-Contiguity spatial weighting (21,634 neighboring relationships). This identifies roads as adjacent if they intersect or overlap. Although it more accurately represents the existing road network, it struggles with handling roads without adjacency (Suryowati et al.,2018).

**3.3 Data labeling and enhancement**

We formalized the restoration quality prediction task as a three-classification task: high-quality, medium-quality, and low-quality. In this study, we labeled the restoration quality categories of each road using a large number of SVIs samples. First, we use a quantitative method to collect SVIs samples based on grid units (50 × 50 meter of each grid) (Dhakal et al.,2000). This method ensures that the samples are as evenly distributed as possible in space, resulting in a total of 2,000 sample points that include the corresponding SVIs (n = 2, 000).

Following this, we developed a restoration quality evaluation platform, drawing inspiration from Zhang et al. (2018). Participants were asked to select the image that best matches the description of the problem in our rating platform (Figure 3). When participants click the Left or Right button, the selected image gains one point correspondingly. When participants selected Neither or Both neither image would gain zero points, or both images would gain one point. Inspired by the work of Celikors and Wells (2022), we selected the best question description for each indicator in the Perceived Restorativeness Scale (PRS) - 11 questionnaire for our study. The PRS-11 questionnaire consists of 11 questions, with two of them describing the scope and three each for Being Away, Coherence, and Fascination (Pasini et al., 2014). Table S2 shows the specific illustration and definition of each question.

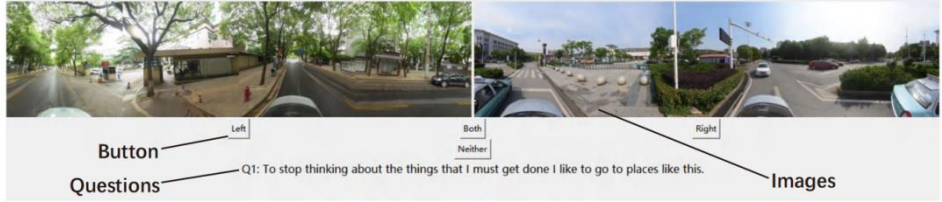

Figure 3: Interface of urban restoration evaluation platform.

During a week-long online survey, we gathered evaluation results from 120 participants, composed of 70 women (Average age = 24.658) and 50 men (Average age = 26.059). The ethical aspects of the experiment were reviewed and approved by our university's institutional review board. After 20 rounds of evaluation (each image was evaluated at least 20 times), we used the Trueskill method to calculate the average score of the four indicators, representing the comprehensive restoration quality of the scene. The Trueskill method is an evaluation and ranking algorithm based on probability theory and statistical theory, suitable for the rating task of each street view image in this study (Herbrich et al.,2006).

Finally, we used the *Jenks Natural Breaks* (Jenks,1967) method to classify the restoration quality of each sample image. *Jenks Natural Breaks* is a data clustering method designed to optimize the arrangement of continuous values into different categories. This method is applicable to the urban restoration quality rating for each individual. We spatially concatenated the classified samples to label the restorative quality categories of units according to the OSM road buffer. As shown in Table 2, we labeled 1,151 roads, with an overall labeling rate of 22.6%, and the proportions of high and medium restoration quality labels remain balanced.

Table 2
Restorative quality label quantity statistics.

| Labels | Number | Proportion (%) |
|---|---|---|
| Low-quality | 434 | 0.377 |
| Medium-quality | 359 | 0.312 |
| High-quality | 322 | 0.279 |
| **Total** | **1151** | |

## 3.4 Model training and evaluation

GCN has an excellent ability to extract graph features, making it suitable for semi-supervised learning tasks and requiring fewer iterations to converge (Kipf and Welling,2016; Zhang et al.,2019). Our research can be categorized as anode classification task. We generate a road graph $A$ based on the adjacency matrix and consider each road as anode $N$, with dimension city features $D$ as the initial matrix $X$. We generate the output at the node level $Z$ (a size matrix of characteristics $N \times F$), where $F$ is the number of characteristics of the output for each node. The GCN propagation rule is as follows 1:

$$f(H^{(l)}, A) = \sigma \left( \hat{D}^{-\frac{1}{2}} \hat{A} \hat{D}^{-\frac{1}{2}} H^{(l)} W^{(l)} \right) \tag{1}$$

where $W^{(l)}$ is a weight matrix for the $l$th layer of the neural network, and is a nonlinear activation function ($ReLU$). $\hat{A} = A + I$, where $I$ is the identity matrix. $\hat{D}$ is the diagonal node degree matrix of $\hat{A}$. Not only the features of the nodes are considered. The adjacency matrix is also regularized. The training process of the model is defined in the following way, using cross-entropy as a loss function quantifying the difference between the probability distributions, where $p$ is the true label and $q$ is the predicted label.

$$cross\_entropy = -\sum_{k=1}^{N} \left( p_k \times \log q_k \right) \tag{2}$$

In graph neural networks, layer number indicates the maximum distance between node features. However, stacking too many layers can result in losing the ability to extract local features (Xu et al., 2022b). In our study, the GCN model was configured with two layers, where a 2-layer GCN can obtain feature information from neighbors and their neighbors. The hidden channels were set to 64 and 32 or 32 and 16. The cross-entropy loss function (2) and the propagation rule equations (1) were used to train the weight matrix $W$. The graph features are transformed from 279 to 3 dimensions using the $Softmax$ function (3), which represents the probability that each node will be classified into one of the three classes.

$$Z = f(X, A) = soft\max \left( \hat{A} \operatorname{ReLU} \left( \hat{A} X W^{(0)} \right) W^{(1)} \right) \tag{3}$$

In addition, we used the Graph Attention Network (GAT) (Veličković et al.,2017) and GraphSAGE

(Hamilton et al., 2017) to assess performance. Unlike the GCN model, the GAT model considers weight parameters between nodes. GraphSAGE only considers the influence of neighboring nodes on the target node, resulting in a faster computational speed. The accuracy score and F1 score (Hossin and Sulaiman,2015) were used to assess the model performance.

## 4. Results

### 4.1 Spatially dependent model performance and prediction results

We carried out 13 experiments using various models, as depicted in Table 3, to measure the average accuracy and F1 score after running the model 10 times. These included intra-group comparisons with four graph neural network classification models using two different spatial weights, and inter-group comparisons with five traditional classification methods, which did not consider spatial weights. The four graph neural network classification models (GCN1, GCN2, GraphSAGE, GAT) conducted 500 training epochs. Three tree-based classification models, Gradient Boosting Decision Trees (GBDT), Random Forest (RF), and Decision Tree (DT), were subjected to 5-fold cross-validation while maintaining consistent parameter settings. Support Vector Machines (SVM) and Multilayer Perceptron (MLP) were used as supplementary methods. Averaged accuracy was calculated after running each model 10 times, and detailed parameter settings can be found in our code depository.

Observing the results of the GNNs, the GAT model exhibited the best classification performance (Acc = 0.735, F1 = 0.732), followed by GraphSAGE (Acc = 0.696, F1 = 0.692) and GCN2 (Acc = 0.635, F1 = 0.622). We noticed that different spatial weights had a significant impact on the model results. For example, in GCN2, using K-Nearest neighbors (Acc = 0.635, F1 = 0.622) yielded better results than using Queen-Contiguity neighbors (Acc = 0.594, F1 = 0.591) within the same timeframe. In general, among the graph neural network models, using K-Nearest neighbors as spatial weights yielded the best classification performance. However, among traditional methods, GBDT achieved the best classification performance (Acc = 0.597, F1 = 0.442), followed by RF (Acc = 0.568, F1 = 0.288) and SVM (Acc = 0.562, F1 = 0.285). It is evident that GNNs, due to their consideration of the current street environment and neighboring street environments, achieve a higher classification accuracy compared to traditional methods with the same data input. This is consistent with the conclusions of previous studies (Zhang et al.,2023; Liu et al.,2023; Liu and Biljecki,2022).

Table 3
Model prediction performance (Epoch=500).

| Model | Weight | Accuracy↑ | F1 Score↑ | Time |
|---|---|---|---|---|
| GCN1 | K-Nearest | 0.613 | 0.610 | 6.130s |
|  | Queen-Contiguity | 0.602 | 0.600 | 6.132s |
| GCN2 | K-Nearest | 0.635 | 0.622 | **6.060s** |
|  | Queen-Contiguity | 0.594 | 0.591 | 6.120s |
| GraphSAGE | K-Nearest | 0.696 | 0.692 | 8.000s |
|  | Queen-Contiguity | 0.613 | 0.609 | 8.030s |
| **GAT** | **K-Nearest** | **0.735** | **0.732** | 23.700s |
|  | Queen-Contiguity | 0.692 | 0.690 | 25.120s |
| RF | - | 0.568 | 0.288 | 31.200s |
| DF | - | 0.486 | 0.367 | 0.781s |
| MLP | - | 0.518 | 0.423 | 16.900s |
| SVM | - | 0.562 | 0.285 | 0.620s |
| GBDT | - | 0.597 | 0.442 | 80.000s |

*Note: GCN1 and GCN2: the hidden channels are 64 and 32, 32 and 16 respectively.*

The GAT model's classification performance was quite satisfactory compared to other models, as shown in the confusion matrix and the T-SNE results (Figure 4). From the confusion matrix, it can be observed that the GAT model can identify all three categories of quality of urban restoration obviously. However, there were some roads in each category that were incorrectly classified as other categories. For example, when predicting high-quality, 18 roads were incorrectly classified as low-quality. We believe this might be due to data imbalance, as medium and low-quality data have a high frequency. Furthermore, we investigated whether the model could effectively learn classification features visually. By reducing the dimension of the output layer of the GAT model, we found that the model has sufficient ability to classify restoration quality based on city features. The following studies are therefore based on the GAT model.

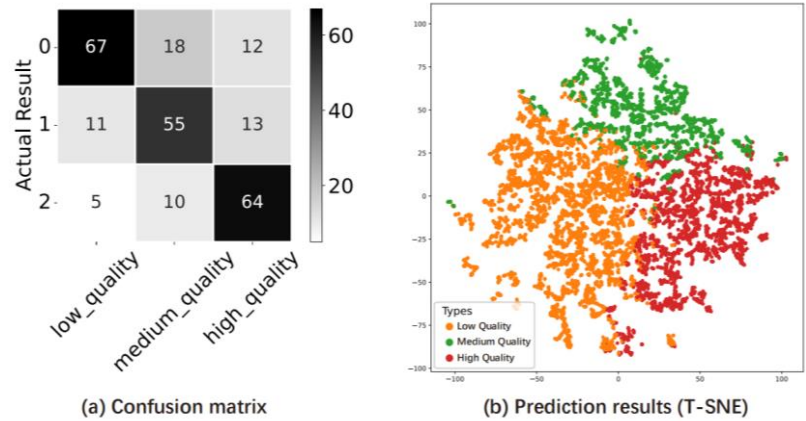

Figure 4: Analysis of GAT model prediction results (spatial weight: K-Nearest, epoch=500). (a) Confusion matrix of model prediction results. (b) T-SNE to reduce the dimension of the output layer of the GAT model.

**4.2 Spatial distribution of restorative perception of urban streets**

As shown in Figure 5a, we mapped the predicted results using the GAT model (highly saturated lines indicate high restoration quality, otherwise the opposite). Within the third ring road of Wuhan, there were 1,344 high-quality restoration roads, 1,420 medium-quality restoration roads, and 2,311 low-quality restoration roads. High-quality restoration spaces have shown an aggregated pattern throughout the city, which may be related to the predictive model we used. GAT models consider not only their own features but also the features of neighboring nodes.

High-quality restoration spaces were most prominent in 1, 2, 3, 4, and 5 (Figure 5a). Among them, 1 is the largest freshwater lake in Wuhan, namely Donghu Lake, with charming waterfront spaces. Waterfronts exhibited more significant restoration abilities, producing a cooling sensation in summer and providing a comfortable environment for sightseeing (Burmil et al.,1999). Interestingly, other high-quality restoration spaces were closely related to surrounding parks or green spaces, such as Zhongshan Park around 2, Hanyang Jiangtan Park around 3, and Houxianghe Park around 4. Urban parks or green areas provide abundant natural resources for attention restoration, and the benefits of interacting with nature are evident (Dadvand et al.,2015; Engemann et al.,2019). There were also means that urban greenery infrastructure had impacted the spatial structure of the neighboring streets. Surprisingly, although located in a high-density residential area, 5 was still predicted to have high restoration quality. We believe this is related to the surrounding shopping centers and historic districts, which have been proven to promote attention restoration (Staats et al.,2016; Fornara et al.,2009). Furthermore, we found that low-quality restoration spaces were mainly concentrated in residential areas, such as 6, 7, and 8. Residential areas occupy the largest proportion of land use in Wuhan, which is the most popular city in central China. However, monotonous residential spaces can easily feel boring, and tall buildings reduce the view distance, affecting the quality of restoration (Lindal and Hartig,2013; Zhang et al.,2018).

Based on the predicted results, we categorized the street scenes according to different restoration qualities (Figure 5b). We found that low-quality spaces lacked green vegetation and had high building densities. In medium-quality spaces, there was a higher presence of shrubs, which, to some extent, increased the restoration quality of the space. In high-quality spaces, the proportion of comfortable roads and the abundance of natural urban landscapes added to the attractiveness and charm of the space.

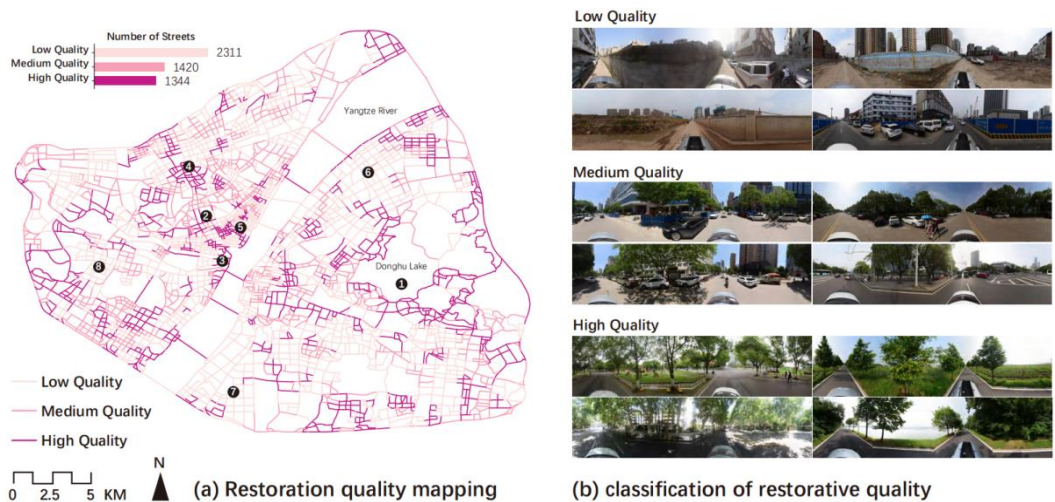

**Figure 5:** Mapping distribution of Wuhan city spatial restoration quality. (a) Restoration quality mapping in Wuhan. (b) Classification of restoration quality. Source of the map: © OSM contributors.

**4.3 Relationship between spatial structure and restoration quality**

As mentioned above, no studies have confirmed the role of spatial structure in restorative environments. We performed ablation experiments based on the GAT model, as shown in Table 4. It can be observed that in Experiment 1, considering all three classes of urban features together, the best classification performance was obtained (Acc = 0.735, F1 = 0.732). However, in Experiment 2, when we removed the spatial-features only, the classification performance decreased significantly (Acc = 0.677, F1 = 0.674).

To further confirm the impact of spatial-features on the prediction results, experiments 3 and 4 were conducted to remove the perceptual-features and socioeconomic- features, respectively. The final results confirmed that spatial-features affect the classification performance of the model (Experiment 3: Acc = 0.704, F1 = 0.699; Experiment 4: Acc = 0.710, F1 = 0.705). Experiment 5, which only considered spatial-features, also exhibited good classification performance in terms of accuracy (Acc = 0.612, F1 = 0.610). In conclusion, the special features increased the prediction accuracy and significantly impacted model performance, which suggests that the spatial structure had a significant impact on human restoration.

**Table 4**
Ablation experiment results based on GAT model.

| Features | Experiment 1 | Experiment 2 | Experiment 3 | Experiment 4 | Experiment 5 |
|---|---|---|---|---|---|
| Socioeconomic-features | ○ | ○ | ○ | × | × |
| Perception-features | ○ | ○ | × | ○ | × |
| Spatial-features | ○ | × | ○ | ○ | ○ |
| **Accuracy** ↑ | 0.735 | 0.677 | 0.704 | 0.710 | 0.612 |
| **F1 Score** ↑ | 0.732 | 0.674 | 0.699 | 0.705 | 0.610 |

Note: ○ stands for retained, and × stands for removed.

To further explore the impact of spatial structure on the quality of urban restoration, we constructed a spatial structure graph containing urban entities, following the method used in Section 3.1. As shown in Figure 6, we first computed for all scenes and generated a spatial structure graph containing all streets in Wuhan (a). Then, based on predicted results for different restoration qualities of urban spaces, three spatial structure graphs were computed (b, c, d). We used node degree centrality to measure the importance of a city entity (node) in the network (Equation 4). The higher degree of a node with a higher centrality degree indicated that the node is more important in the network (Degenne and Forsé,1999). The n represents the number of nodes and $N_{degree}$ represents the degree of that node.

$$Degree\_centrality = \frac{N_{degree}}{n-1} \qquad (4)$$

We calculated the top five entities with the highest degree centrality for each spatial structure graph. It can be observed that building entities have the highest degree centrality in graphs a, b, c, and d (Figure 6). We considered this a normal result since 70% of the study area was covered by buildings. In the overall spatial structure graph (a), the top five entities in terms of centrality were Building (0.883), People (0.866), Car (0.853), Road (0.850) and Tree (0.841).

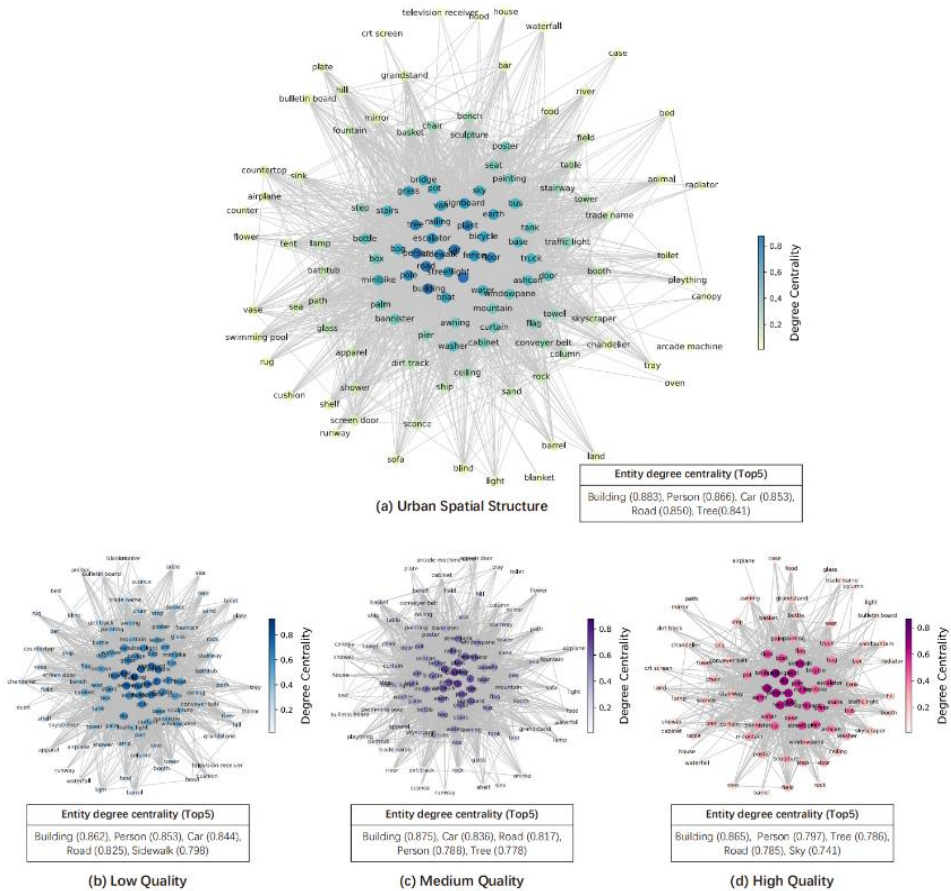

Figure 6: Spatial entity structure graph. (a) A spatial structure graph containing all streets in Wuhan. (b) A low-quality spatial structure graph. (c) A medium-quality spatial structure graph. (d) A high-quality spatial structure graph.

Additionally, we found evidently differences in the centrality of the top five entities between different restoration qualities. Specifically, in low-quality spaces (b), non-natural entities played an important role in the graph, such as Road (0.825) and Sidewalk (0.8798). Conversely, in high-quality spaces (d), some natural entities started to emerge, such as Tree (0.786) and the Sky (0.741). In medium-quality spaces, they were similar to low-quality ones and the only difference was that spaces included natural elements, such as Tree (0.778). In conclusion, spatial structure graphs of different restoration qualities exhibited apparently structural differences, especially among the top five entities in terms of centrality. This further evidence provided a strong explanation for the effect of spatial structure on restoration quality.

**4.4 Spatial structural distinct of high restoration quality**

It is worth noting that in the same restoration quality environment, there were different visual contents due to geographical differences. We used the *K-Means* clustering method to reveal the spatial structure distinct of 1,344 roads predicted as high-quality restoration (Jain, 2010). As shown in Figure 7a, the solution is represented by the average silhouette of 20-class, and the best classification effect is obtained by the solution of 4 clusters. As shown in Figure 7b, the road contents of each cluster are very similar. Ironically, using traditional methods such as semantic segmentation or object counting, spatial heterogeneity is difficult to detect. For example, Cluster_0 and Cluster_ 1 have similar contents, both have almost 37.5% tree percentage, but Cluster_0 is more in residential or commercial areas, while Cluster_ 1 is more in urban parks or waterfront areas.

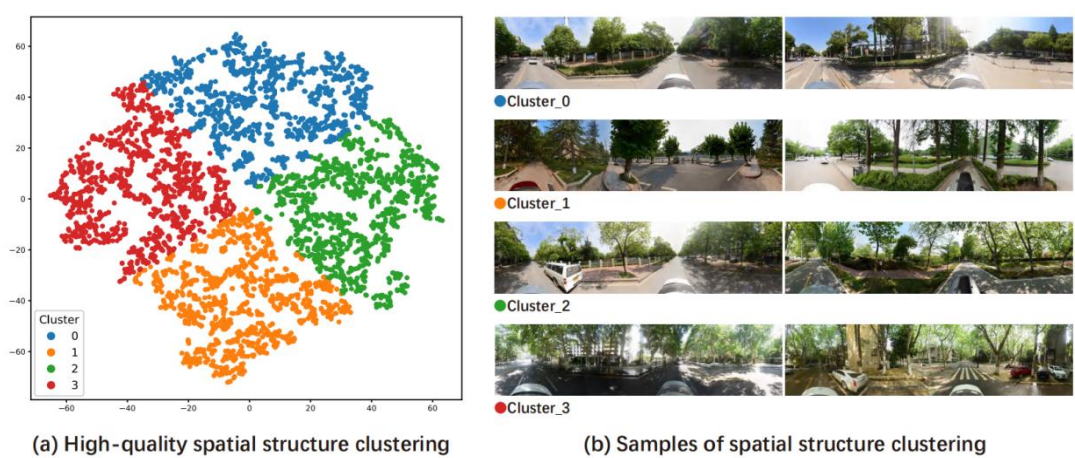

(a) High-quality spatial structure clustering    (b) Samples of spatial structure clustering

Figure 7: High-quality restoration spatial structural differences analysis. (a) High-quality spatial structure clustering plot. (b) Sample of spatial clustering.

Based on the clustering results, we developed four cluster spatial structure graphs and used degree centrality to measure the importance of the entities (Top 10). As demonstrated in Table 5, the spatial structure of Cluster_0 and Cluster_2 is most similar. The first five entities with the highest degree centrality are consistent (Building, Sidewalk, Person, Plant, Road), although the order is different. It is worth noting that the importance of Sky in Cluster_0 is much higher than that in Cluster_2, which means that more blue space can be felt in Cluster_0, and blue, as a crucial urban infrastructure, can bring rich health benefits (Smith et al.,2021). The spatial structure of Cluster_ 1 is the most special, the first entity is Person. The building entity is in the fifth. According to our observation, Cluster_ 1 is mostly in urban parks or waterfront areas, which often gather a large number of people and are far away from high-density urban building space (Nordh et al.,2009; Brancato et al.,2022). The spatial features displayed by Cluster_3 is mostly in the city's tree-lined avenues, and the second entity is Tree. Rich vegetation can provide shade, and attract people to walk or ride, and tree-lined avenues area typical aesthetically pleasing urban landscape (Li et al.,2018). In summary, spatial graphs were more accurate than traditional methods for revealing the distinct of spatial structures, and more spatial attribute information can be extracted, which improves the comprehensibility of environmental restoration.

**Table 5**
Top 10-degree centrality entities of four clusters graph.

| Clusters | Entities degree centrality (Top 10) |
| --- | --- |
| Cluster_0 | Building, Sidewalk, Person, Plant, Road, Sky, Car, Tree, Grass, Earth |
| Cluster_1 | Person, Car, Sidewalk, Road, Building, Sky, Plant, Tree, Pole, Earth |
| Cluster_2 | Building, Person, Sidewalk, Plant, Road, Car, Tree, Sky, Earth, Floor |
| Cluster_3 | Building, Tree, Sidewalk, Person, Plant, Sky, Road, Car, Pole, Railing |

## 5. Discussion

This study introduces a spatial-dependent GNNs approach to predict urban restoration quality and reveal the relation between spatial structure and restoration quality. By embedding spatial unit graphs through multiple sequential SVIs, the influence of spatial structure on restorative quality and the structural heterogeneity of restorative space are explored. Our findings suggest that the spatially dependent GNNs, through learning from the fusion of various features and geographic relationships, not only take into account the environment of specific locations but also provide a holistic perspective. This approach facilitates a comprehensive understanding of restoration characteristics at an urban scale, maximizing the consideration of interdependency between spaces. Therefore, our study fills a gap in understanding how spatial structure influences restoration quality.

We have found that spatial structure strongly determines the restorative quality of urban environments. In spaces with low restoration quality, the importance of non-natural entities is significantly higher, while in spaces with high restoration quality, the opposite is true, with natural entities becoming more important. This is consistent with previous research, which shows a positive correlation between natural elements and restoration, together with various health benefits (Capaldi et al.,2014; Schertz and Berman, 2019). Furthermore, spaces with the same restoration quality exhibit different spatial structures due to entity relationships in continuous spaces are different. Through cluster analysis, we identified four patterns of high-quality restoration spatial structures. The verification of the accuracy of spatial graphs in highlighting differences in spatial structure enhances the interpretability of restoration spaces, which is crucial for guiding restoration space design across different urban areas.

In addition to the influence of spatial structure, our research has also revealed the distribution characteristics of high-quality restoration spaces in cities. Notably, urban waterfront spaces emerge conspicuously. Water views are considered positive restorative visual components, and people tend to walk or cycle in areas with abundant water views to reduce stress (Massoni et al.,2018). Furthermore, the characteristics of natural water can seamlessly integrate into the surrounding natural landscapes, enhancing aesthetic experiences and restoring attentional resources (Markevych et al., 2017; Roe et al., 2019). Therefore, waterfront landscape types should be given priority in urban design and redevelopment. Moreover, urban green spaces are closely associated with high restoration quality, and numerous studies have demonstrated the psychological benefits of urban green spaces, such as urban parks (Nordh et al.,2011,2009). This research thus provides further evidence supporting the restorative potential of cities.

Additionally, while SVIs are emerging as an important data source in urban research (Tang and Long,2019; Biljecki and Ito,2021), there have been limited studies using it to investigate spatial restoration quality on an urban scale. SVIs demonstrate significant advantages in research. First, it has wide coverage, fast updates, and precise geographic coordinates. Second, it has a wealth of visual and spatial information, and when used to predict human perception, there is a minimal amount of data bias (Kang et al.,2023; Zhao et al.,2023; Ma et al.,2023). In this study, to predict urban restoration quality, a large number of SVIs were used to extract visual features and human perceptions, and embed the spatial structure in sequential scenes. The results proved that SVIs can accurately predict urban restoration quality and demonstrated the advantages of representing the spatial structure.

Simultaneously, this study has some limitations. First, urban environments are dynamic and influenced by factors such as human activities, urban functions, and traffic conditions (Kaplan and Herbert,1987;

Quercia et al., 2014). Complex features, such as sound and temperature, can affect the restoration quality (Hartig et al., 2007; Qi et al., 2022; Ratcliffe, 2021). In the future, it is possible to expand to a broader range of dimensions by integrating multi-modal data or digital environments that incorporate these elements that potentially influence the quality of environmental restoration. Second, the city features captured by SVI data only reflect specific moments in urban scenes, thus exhibiting a temporal lag. Also, SVI is primarily captured from a driving perspective, lacking the perception of content from a human perspective (Biljecki and Ito, 2021). Finally, in a restoration environment, the structurally simple PRS-11 may weaken the accuracy of the assessment of environmental restoration quality. Future research could consider incorporating more diverse attention restoration questionnaires and questions to improve the precision of the results.

## 6. Conclusion

The long-standing discussion on the relationship between restorative quality and the physical environment, however, lacked research on the impacts caused by diverse spatial structures and scarce more efficient ways for measuring on an urban scale. Our study proposed a spatial-dependent GNNs approach for solving these questions, which includes two types of graphs: street and city levels. This study made three contributions. First, we proposed a spatial-dependent prediction method for measuring urban restoration quality by capturing road topology relationships using graph neural networks and aggregating contextual features of cities as a city-level graph. Second, we used a novel graph approach to reveal spatial structure effects among different restoration qualities by capturing the intrinsic and extrinsic relationships between entities through sequential SVIs. Third, we further uncovered the spatial structural distinct among high-quality restoration spaces caused by geographic location differences, thereby enhancing the understandability of restorative spatial features. Overall, this study provides insight into healthy city construction, improves the interpretability of urban restoration spaces, and can be applied further to the design of healthy medium-scale spaces, such as communities or parks.